\documentclass[sigconf]{acmart}
\usepackage{multirow}
\usepackage{titlesec}
\titleformat{\subsubsection}[runin]{\normalfont\normalsize\bfseries}{\thesubsubsection}{1em}{}[]

\titlespacing{\subsubsection}{0pt}{1.0ex plus 1.0ex minus .2ex}{1.5ex plus .2ex}

\usepackage{enumitem}
\setlist[itemize]{align=parleft,left=8pt..1.8em}

\AtBeginDocument{%
	}

\copyrightyear{2023} 
\acmYear{2023} 
\setcopyright{rightsretained} 
\acmConference[CIKM '23]{Proceedings of the 32nd ACM International Conference on Information and Knowledge Management}{October 21--25, 2023}{Birmingham, United Kingdom}
\acmBooktitle{Proceedings of the 32nd ACM International Conference on Information and Knowledge Management (CIKM '23), October 21--25, 2023, Birmingham, United Kingdom}\acmDOI{10.1145/3583780.3615126}
\acmISBN{979-8-4007-0124-5/23/10}

\begin{document}
	\title{ 
		Ontology Enrichment from Texts: A Biomedical Dataset for Concept Discovery and Placement}
	
	\author{Hang Dong}
	\affiliation{%
		\institution{University of Oxford}
		\country{}
	}
	\email{hang.dong@cs.ox.ac.uk}
	
	\author{Jiaoyan Chen}
	\affiliation{%
		\institution{The University of Manchester \& University of Oxford}
		\country{}
	}
	\email{jiaoyan.chen@manchester.ac.uk}
	
	\author{Yuan He}
	\affiliation{%
		\institution{University of Oxford}
		\country{}
	}
	\email{yuan.he@cs.ox.ac.uk}
	
	\author{Ian Horrocks}
	\affiliation{%
		\institution{University of Oxford}
		\country{}
	}
	\email{ian.horrocks@cs.ox.ac.uk}
	
	\renewcommand{\shortauthors}{Hang Dong, Jiaoyan Chen, Yuan He, \& Ian Horrocks}
	
	\begin{abstract}
		\noindent\textit{Mentions of new concepts appear regularly in texts and require automated approaches to harvest and place them into Knowledge Bases (KB), e.g., ontologies and taxonomies. Existing datasets suffer from three issues, (i) mostly assuming that a new concept is pre-discovered and cannot support out-of-KB mention discovery; (ii) only using the concept label as the input along with the KB and thus lacking the contexts of a concept label; and (iii) mostly focusing on concept placement w.r.t a taxonomy of atomic concepts, instead of complex concepts, i.e., with logical operators. To address these issues, we propose a new benchmark, adapting MedMentions dataset (PubMed abstracts) with SNOMED CT versions in 2014 and 2017 under the Diseases sub-category and the broader categories of Clinical finding, Procedure, and Pharmaceutical / biologic product. We provide usage on the evaluation with the dataset for out-of-KB mention discovery and concept placement, adapting recent Large Language Model based methods.\footnote{The dataset, data construction scripts, and baseline implementation are available at \url{https://zenodo.org/record/8228005} and \url{https://github.com/KRR-Oxford/OET}.}}
	\end{abstract}

	
	
	\maketitle

	\section{INTRODUCTION}
	\label{sec:intro}
	Identifying new concepts and placing them into Knowledge Bases (KBs, e.g., ontologies and taxonomies) from texts such as a vast amount of publications is a key application of KB construction and AI for scientific discovery \cite{hastings2023ai}. Emerging concepts are particularly common in the biomedical domain and KB can easily be outdated. For example, new variants of SARS-CoV-2 have kept emerging since 2020; ``Curry-Jones syndrome'' was not added to SNOMED CT ontology \cite{donnelly2006snomed} until 2017.
	
	Existing datasets on using texts to enrich ontologies are relevant to several tasks, but each of them only reflects a part of the whole picture. In \textit{Taxonomy Completion} \cite{Shen2020TaxoExpan,jurgens-pilehvar-2016-semeval}, a pre-specified out-of-KB (a.k.a. NIL) concept is used to enrich a taxonomy. In \textit{Ontology Extension} \cite{glauer2022interpretable}, this is extended into description logic based ontologies, which include complex concepts which can be considered \cite{chen2023contextual,yuan2023ontolama} but not yet for the existing datasets \cite{glauer2022interpretable}. In \textit{Concept Post-coordination} \cite{kate2020,castell2023postcoordination}, an out-of-KB concept is defined with several existing concepts and attributes, i.e., placed under a complex concept. Datasets from all the three tasks above assume that the input concept term is already specified and non-contextual (e.g., without contexts in a corpus), which does not reflect the real-world situation. In \textit{Out-of-KB Mention and Entity Discovery} \cite{Iurshina2022nilk,heist2023nastylinker,dong2023NIL}, out-of-KB mentions and their clustering are discovered from texts, but their placement in KBs has not been fully investigated.
	\begin{table*}[ht]
		\small
		\caption{Comparison of relevant datasets and tasks on KB (e.g., ontology and taxonomy) enrichment from texts. \textit{NIL Discovery} denotes whether the task can support discovering out-of-KB mentions (\textit{cf.} in-KB mentions). \textit{Contextual Term} denotes whether the input term has a context window in a text corpus. \textit{Concept Placement} denotes whether the task finally places (or can be used to place) the term in the KB. \textit{Complex Concepts} denote whether the placement position in the KB includes complex concepts. The asterisk (*) denotes that only data construction scripts are available instead of the dataset itself. }\label{dataset-summary}
		\vspace{-0.2cm}
		\begin{tabular}{p{5.2cm}p{2.4cm}|llll}
			\cline{1-6}
			Datasets (with public access link in citations) & Task                & NIL Discovery & Contextual Term  & Concept Placement  & Complex Concepts \\
			\cline{1-6}
			MAG
			, WordNet 
			\cite{zhang2021TMN,Shen2020TaxoExpan,jurgens-pilehvar-2016-semeval}; OSConcepts, DroneTaxo, MeSH, SemEval 
			\cite{Zeng2021,Wang2022QEN}                     & Taxonomy Completion                              & No                 & No                & Yes & No                  \\ 
			\cline{1-6}
			ChEBI$_{500}$, ChEBI$^+_{500}$ \cite{glauer2022interpretable} & Ontology Extension
			& No                 & No & Yes & No                  \\ \cline{1-6}
			SNOMED CT (English, manual, small-scale) \cite{kate2020}; SNOMED CT (Spanish, automated) \cite{castell2023postcoordination}$^*$                     & Concept Post-coordination                        &   No                 & No   & Yes               & Yes                  \\ 
			\cline{1-6}
			NILK \cite{Iurshina2022nilk}; ShARe/CLEF 2013 \cite{suominen2013shareclef}; CLEF HIPE 2020 \cite{ehrmann2020extended}; EDIN \cite{kassner2022edin}; NEEL 2015-2016 \cite{rizzo2015neel,rizzo2016neel}
			& Out-of-KB Mention and Entity Discovery &   Yes              & Yes  & No                & No                   \\
			\hline\hline 
			MedMentions-SNOMED-CT-14 (-CPP, -Disease) [\textbf{this work}]                     & Concept Discovery and Placement                             &  Yes                & Yes               & Yes  & Yes                
			\\
			\cline{1-6}
		\end{tabular}
	\end{table*}
	
	In this study, we propose a new benchmark for new entity discovery and placement, supporting two sequential tasks: (i) \textit{Out-of-KB Mention and Entity Discovery}: identifying new mentions of concepts from texts which are not included in a KB; (ii) \textit{Concept Placement}: given a \textit{new} entity expressed as a mention in the text, placing it into a KB, either an ontology with complex concepts or a taxonomy with only atomic concepts. Our new dataset and task setting are different from previous work in terms of the characteristics below:
	
	\begin{itemize}
		\item \textbf{Out-of-KB or NIL discovery}: inclusion of \textit{out-of-KB mentions} from texts to support their concept discovery and placement.
		\item \textbf{Contextual terms}: inclusion of \textit{contexts} for mentions, distinct from only using concept labels as in the previous work.
		\item \textbf{Complex concepts}: placement of concepts under logic-\\equipped \textit{complex concepts}, instead of atomic concepts alone.
	\end{itemize}
	
	More specifically, the study uses a SNOMED CT subset as the ontology, and the time difference of two versions (in 2014 and 2017) to synthesise new entities, then uses MedMentions Entity Linking dataset \cite{Mohan2019medmentions} (from PubMed abstracts to UMLS) to construct in-KB and out-of-KB mentions. The study further introduces the usage of the data with evaluation for out-of-KB mention discovery and concept placement. We provide benchmarking results adapting rule-based and BERT-based Entity Linking \cite{dong2023NIL,wu2020blink} and prompting with GPT-3.5-turbo. Results show that the dataset well differentiates the performance between rule-based and BERT-based methods, and the
	Pre-trained and Large Language Model (LLM) based methods are still yet to achieve satisfying results.
	
	\section{RELATED WORK}
	The representative datasets are summarised in Table \ref{dataset-summary} based on the four tasks introduced in Section \ref{sec:intro}. We only list the public and accessible datasets. We next discuss the related work of each task.
	
	\noindent\textbf{Taxonomy Completion and Ontology Extension}. Studies in taxonomy completion \cite{Wang2022QEN,Zeng2021,zhang2021TMN,Shen2020TaxoExpan,jurgens-pilehvar-2016-semeval} and ontology extension \cite{glauer2022interpretable} aim to enrich KB using the concept labels and the concept graph structure. However, the studies usually assume that the new term (or concept label) is pre-discovered, which is not the case in the real-world scenario, where new mentions of concepts need to be discovered from corpora. Also, from the perspectives of OWL (Web Ontology Language) \cite{GRAU2008OWL2,baader_horrocks_lutz_sattler_2017chap8}, 
	most of these studies focus only on atomic concepts and do not place the new concept under a \textit{complex concept}, e.g., with existential restrictions used in SNOMED CT (e.g., \cite{Liu2020placement} focuses on placement under only atomic concepts in SNOMED CT). Also, datasets in both tasks use concept terms as input and do not consider contexts. 
	
	\noindent\textbf{Concept Post-coordination}. The studies aim to place a new concept by describing it with existing concepts and attributes in the ontology \cite{kate2020,castell2023postcoordination}. Dataset construction steps in both works \cite{kate2020,castell2023postcoordination} assume that the new concepts or terms are pre-discovered and without context windows from a corpus. 
	
	\noindent\textbf{Out-of-KB Mention and Entity Discovery}. The studies aim to discover new mentions from texts, w.r.t. to a KB \cite{dong2023NIL,suominen2013shareclef} and group them into entity clusters \cite{Iurshina2022nilk,kassner2022edin,rizzo2017NEEL,ji2011TAC}. There is a growth of datasets in this area recently, constructed through Manual Labelling, KB pruning, and/or KB versioning \cite{dong2023NIL}. The studies, however, do not place the newly discovered entities into a KB. 
	
	In this work, we present dataset construction for \textit{\textbf{Concept Discovery and Placement}} to support a comprehensive set of characteristics (Table \ref{dataset-summary}), with usage for benchmarking, e.g., with Pre-trained and Large Language Models. 
	
	\section{PROBLEM DEFINITION}
	
	The task of \textit{Concept Discovery and Placement} inputs contextual, in-KB and out-of-KB mentions in a corpus and a KB (more formally as an OWL ontology \cite{GRAU2008OWL2,baader_horrocks_lutz_sattler_2017chap8}) and outputs an enriched KB where each out-of-KB mention is inserted into a \textit{directed edge}, i.e., $<\texttt{parent, child}>$, of the KB, when the out-of-KB mention is the child of the \texttt{parent} and the parent of the \texttt{child}. The \texttt{child} is considered NULL when the mention corresponds to a leaf concept. The \texttt{parent} can be a \textit{complex concept}. 
	
	\begin{figure*}[ht]
		\includegraphics[width=\textwidth]{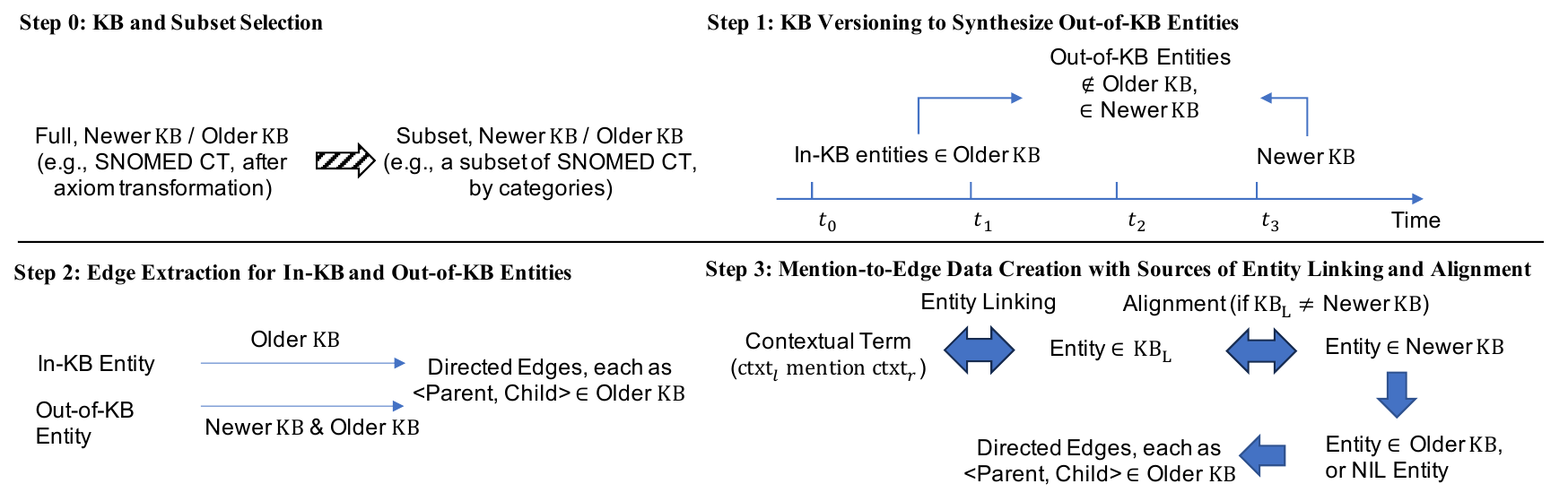}
		\caption{Data construction pipeline: KB and Subset Selection, KB Versioning, Edge Extraction, and Mention-Edge Data Creation.}\label{data-construction-steps} 
		\vspace{-3mm}
	\end{figure*} 
	
	Several key definitions are as follows. Formally, an \textit{OWL ontology} is a Description Logic KB that contains a set of 
	axioms \cite{GRAU2008OWL2,baader_horrocks_lutz_sattler_2017chap8}. We focus on the TBox (or the terminology part) in an ontology, which mainly consists of General Concept Inclusion axioms of the form $A \sqsubseteq B$, where $A$ (and $B$) are either atomic or complex concepts \cite{baader_horrocks_lutz_sattler_2017chap2}. A TBox can be reduced to a \textit{taxonomy} (or a subsumption hierarchy) after the classification process \cite{baader_horrocks_lutz_sattler_2017chap2}. Borrowing the definition of taxonomy in \cite{Jiang2022taxoenrich,Shen2020TaxoExpan}, we use a simple definition of ontology as a set of concepts and directed edges, where both can be atomic or complex. \textit{Directed Edges} are edges in an ontology (or a taxonomy \cite{Wang2022QEN}) which contain a direct parent and a direct child. \textit{Complex edges} are edges which have a complex concept as the parent\footnote{It is not likely to have complex concepts as direct (asserted) children in an ontology.}. \textit{Complex concepts} mean concepts that involve at least one logical operators, e.g., negation ($\neg$), conjunction ($\sqcap$), disjunction ($\sqcup$), existential restriction ($\exists r.C$), universal restriction ($\forall r. C$), etc. \cite{baader_horrocks_lutz_sattler_2017chap2}.\footnote{The ontology language OWL \cite{GRAU2008OWL2}, based on Description Logic, has expressiveness beyond RDF \cite{baader_horrocks_lutz_sattler_2017chap8}; an example complex concept as a parent concept is $Arthritis \sqsubseteq Arthropathy \sqcap \exists hasMorphology.Inflammatory$  \cite{yuan2023ontolama}.}  
	
	An ideal dataset for \textit{Concept Discovery and Placement} requires a real-world text corpus and a large OWL ontology (reducible to a taxonomy), with gold-standard directed edges (possibly complex) for each out-of-KB concept linked to the mentions in the corpus.
	
	\section{DATASET CONSTRUCTION}
	
	\subsubsection*{Step 0: KB and Subset Selection.}
	We consider SNOMED CT \cite{donnelly2006snomed}, one of the most important OWL ontology in the biomedical domain, and choose a subset by selected categories of concepts. We focus on the second level category, \textit{Disease} (disorder) and the first level categories, \textit{Clinical finding}, \textit{Pharmaceutical / biologic product}, and \textit{Procedure}, abbreviated to \textit{CPP} as the initials of the categories. CPP categories have the most important types of complex edges for placement (or post-coordination), according to \citet{kate2020}.
	
	The subset selection has two steps: (i) transforming equivalence axioms into subsumption axioms by outermost conjunctions, e.g., from $A \equiv C_1 \sqcap \exists R.(C_2 \sqcap C_3)$ to $A \sqsubseteq C_1$ and $A \sqsubseteq \exists R.(C_2 \sqcap C_3)$; (ii) pruning the ontology to only keep the selected categories.\footnote{Given that many subsumption relations are in fact described as equivalence axioms with conjunction with concepts from to-be-removed categories, simply using step (ii) without step (i) will result in many broken hierarchies instead of a connected one.}
	
	The implementation is as follows. First, we transform SNOMED CT files\footnote{\url{https://www.nlm.nih.gov/healthit/snomedct/archive.html}} into OWL format with snomed-owl-toolkit\footnote{\url{https://github.com/IHTSDO/snomed-owl-toolkit}}. Then, step (i) is implemented through ontology processing with DeepOnto\footnote{\url{https://github.com/KRR-Oxford/DeepOnto}} \cite{he2023deeponto}; and step (ii) through Protégé\footnote{\url{http://protegeproject.github.io/protege/}} to remove the other categories.
	
	\subsubsection*{Step 1: KB Versioning.}
	We follow a KB versioning strategy \cite{dong2023NIL,Iurshina2022nilk} to synthesise out-of-KB entities for the older KB. The concept gap between the two versions of SNOMED CT subsets (ver 20140901 and 20170301) is considered. The numbers of concepts in the older and the newer sub-KB are 64,900 and 72,595, resp., for the \textit{Disease} sub-category and are 175,895 and 188,988, resp., for \textit{CPP} categories.
	
	\subsubsection*{Step 2: Edge Extraction.}
	We extract the directed edges in the older sub-KB for both in-KB and out-of-KB entities. For in-KB entities, this is achieved by querying all the direct parents and children in the older KB. For out-of-KB entities, this is achieved by querying all the \textit{most} direct, in-KB (older KB) parents and children from the newer sub-KB, given that edges for the out-of-KB entities are not available in the older KB. If the entity is a leaf node, we set the direct child as NULL, as in \citet{zhang2021TMN}. The querying process is based on ontology processing module in DeepOnto \cite{he2023deeponto}.
	
	\subsubsection*{Step 3: Mention-Edge Data Creation.}
	A corpus with mentions linked to entities in SNOMED CT is needed to synthesise contextual mentions and gold-standard edges. Following \citet{dong2023NIL}, we use MedMentions\footnote{\url{https://github.com/chanzuckerberg/MedMentions}} \cite{Mohan2019medmentions}, containing around 5,000 biomedical paper abstracts in PubMed\footnote{\url{https://pubmed.ncbi.nlm.nih.gov/}} where the mentions were manually and exhaustively linked to UMLS \cite{Bodenreider2004umls} (version 2017AA)\footnote{There are also other datasets available, e.g. COMETA \cite{basaldella2020cometa}, which directly links mentions in social media posts on Reddit to SNOMED CT. Still, we consider scientific publications a more reliable source and leave social media for a future study.}. We leverage the alignment of entities in UMLS to obtain  SNOMED CT entities (ver 20170301)\footnote{If mention of one UMLS entity matches multiple SNOMED CT entities, we create a mention row in the data for each SNOMED CT entity.}. Using the output from the previous steps, we can then map each in-KB and out-of-KB mention to edges\footnote{We also filtered the edges to one-hop (including leaf concept to NULL) and two-hop from any paths in the ontology.} in the older sub-KB (SNOMED CT ver 20140901).
	
	We thus create mention-to-edge datasets with the information for each mention, each rendered in a JSON format. The information includes the left and the right contexts of the mention ($\text{ctxt}_l$ and $\text{ctxt}_r$), the mention or concept itself and its SNOMED CT ID, the parent and child in the older SNOMED CT ID (and with expression for complex concepts) and their labels.\footnote{The keys in the JSON format are available in \url{https://github.com/KRR-Oxford/OET}.} We use DeepOnto's verbaliser \cite{he2023deeponto,yuan2023ontolama} to form labels of the complex concepts.
	
	\begin{table*}[ht]
		\small
		\caption{Statistics for datasets for \textit{Concept Discovery and Placement}, for SNOMED CT (ver 20140901, ``S14'') under different categories: ``Disease'' and ``CPP'', i.e., \textit{\underline{C}linical finding}, \textit{\underline{P}rocedure}, and \textit{\underline{P}harmaceutical / biologic product}. A mention-edge pair denotes a mention (in a corpus) and one of its directed edges in the KB. 
			Mentions are from the MedMentions dataset (``MM''). \\ 
			* The numbers of edges are those having one hop (including leaf nodes to NULL) and two hops from any paths in the ontology.
		}\label{data-statistics}
		\begin{tabular}{l|l|l|lll|l}
			\cline{1-7}
			
			& \multicolumn{2}{c}{Ontology: \# all (\# complex) }                                           & \multicolumn{4}{|c}{Corpus: \# Mentions / \# Mention-edge pairs / \# Mention-edge pairs with \textit{complex} edges}                             \\ \hline                       & concepts & edges* & train, in-KB   & valid, in-KB & test, in-KB & out-of-KB \\
			\cline{1-7}
			MM-S14-Disease & 64,900 (824)        & 237,826 (4,997)      & 11,812 / 887,840 / 917  & 4,248 / 383,457 / 203    & 3,970 / 316,319 / 393    & 605 / 1,637 / 13 \\             
			MM-S14-CPP     & 175,895 (2,718)      & 625,994 (19,401)     & 34,704 / 1,398,111 / 9,475 & 11,707 / 548,295 / 4,305     & 11,564 / 478,424 / 4,129     & 1,000 / 2,131 / 22                  \\
			\cline{1-7}
		\end{tabular}
	\end{table*}
	
	Regarding data splitting for the benchmark, for \textit{out-of-KB mention and concept discovery}, the dataset follows the original splits of training, validation, and testing sets from MedMentions; for \textit{concept placement}, the setting is unsupervised for out-of-KB mentions, i.e., training (and validating) with in-KB mentions but testing on out-of-KB mentions (and in-KB mentions)
	
	We provide two formats of the data, \textit{mention-level}, with edges grouped for each mention; and \textit{mention-edge-pair-level}, where each mention-edge pair occupies a row and mentions are repeated if there are multiple edges. Statistics of the datasets are in Table \ref{data-statistics}. 
	
	\section{DATA USAGE FOR BENCHMARKING}
	
	\subsection{Evaluation with the Data}
	\subsubsection*{Metrics for Out-of-KB Mention Discovery} The dataset supports the metrics in \cite{dong2023NIL}, including overall accuracy for all in-KB and out-of-KB mentions ($A$); \textit{out-of-KB} precision ($P_o$), recall ($R_o$), and $F_1$ score ($F_{1_o}$) to measure how well out-of-KB mentions are detected; and \textit{in-KB} precision ($P_{in}$), recall ($R_{in}$, and $F_1$ score ($F_{1_{in}}$).
	
	\subsubsection*{Metrics for Concept Placement} The dataset supports the metrics used in taxonomy completion, to evaluate the ranking of edges for a given mention \cite{zhang2021TMN,Jiang2022taxoenrich,Shen2020TaxoExpan}. The metrics mainly include Precision at $k$ (P@$k$), Recall at $k$ (R@$k$), $F_1$ score at $k$ ($F_1$@$k$), Mean Rank (MR), and Mean Reciprocal Rank (MRR). We report P@$k$ and R@$k$ for different top-$k$ values. 
	
	\subsection{Experimenting with the Data}
	We experiment with the data w.r.t the two tasks using a rule-based method and recent, LLM-based methods.
	
	\begin{table}
		\small
		\caption{Results on out-of-KB mention discovery}\label{out-of-KB-res}
		\begin{tabular}{l|l|lll|lll}
			\cline{1-8}
			MM-S14-Disease & A & $P_o$ & $R_o$ & $F_{1_o}$ & $P_{in}$ & $R_{in}$ & $F_{1_{in}}$    \\
			\cline{1-8}
			Sieve-based & 55.9 & 6.4      & \textbf{47.0}  & 11.2 & \textbf{88.1} & 56.2 & 68.6  \\
			BLINKout       & \textbf{67.2} & \textbf{14.6}      & 17.4     & \textbf{15.9} & 69.0 & \textbf{68.6} & \textbf{68.8}  \\
			\hline\hline
			MM-S14-CPP     & A & $P_o$ & $R_o$ & $F_{1_o}$ & $P_{in}$ & $R_{in}$ & $F_{1_{in}}$    \\
			\cline{1-8}
			Sieve-based & 49.7 & 3.3      & \textbf{59.3}   & 6.3 & \textbf{85.7} & 49.6 & 62.8  \\
			BLINKout       & \textbf{65.9} & \textbf{22.5}          & 32.6      & \textbf{26.6} & 66.8 & \textbf{66.4} & \textbf{66.6}     \\
			\cline{1-8}
		\end{tabular}
	\end{table}
	
	\begin{table}
		\small
		\caption{Results on out-of-KB concept placement}\label{insertion-res}
		\begin{tabular}{l|l|l|l|l}
			\cline{1-5}
			MM-S14-Disease  & P / R @1 & P / R @5 & P / R @10 & P / R @50\\
			\cline{1-5}
			Edge-Bi-encoder &  \textbf{4.5} / \textbf{1.6}         & 6.0 / 11.0         & 5.4 / 19.9 & 2.1 / 38.4  \\
			+GPT-3.5    & 4.3 / 1.6          &    -       &   -      &  -  \\
			\hline\hline
			MM-S14-CPP      & P / R @1 & P / R @5 & P / R @10 & P / R @50 \\
			\cline{1-5}
			Edge-Bi-encoder &  2.2 / 1.0 & 2.2 / 5.2 & 2.0 / 9.4 & 1.4 / 32.4\\ 
			+GPT-3.5        & \textbf{2.5} / \textbf{1.2}          &   -        &   -   &   -   \\ 
			\cline{1-5}
		\end{tabular}
	\end{table}
	
	\subsubsection*{For Out-of-KB Mention Discovery} Existing methods are supervised, i.e., require a certain amount of NIL in the training data. Thus, we split the ``out-of-KB'' mentions in Table \ref{data-statistics} based on the NIL mentions in the original MedMentions data split. The number of training, validation, and testing NIL mentions are 568, 260, and 172, resp., for CPP (in total 1,000 mentions); and 329, 161, and 115, resp., for Disease sub-categories  (in total 605 mentions). For the rule-based method, we use Sieve-based approach, which uses rules designed for biomedical texts and predicts a mention as out-of-KB if no in-KB entity can be linked to \cite{dsouza-ng-2015-sieve-based}. For the LLM-based method, we follow BLINKout \cite{dong2023NIL} to detect out-of-KB mentions from texts adapting a two-step BERT-based approach \cite{wu2020blink}: candidate generation with bi-encoder and candidate selection with cross-encoder. Out-of-KB mentions are discovered through NIL entity representation and classification in the cross-encoder \cite{dong2023NIL}. We used default parameters with top-$k$ value as 50 and domain-specific model, SapBERT \cite{liu2021sap}.
	
	\subsubsection*{Results on Out-of-KB Mention Discovery} Table \ref{out-of-KB-res} shows that BLINKout performs much better than the Sieve-based approach in terms of the overall accuracy and out-of-KB $F_1$ scores. However, it is still challenging to achieve satisfying performance to identify out-of-KB mentions (with out-of-KB $F_{1}$ between 15\% and 30\%).
	
	\subsubsection*{For Concept Placement} We use the mention-edge pairs (see Table \ref{data-statistics}) to train and validate a model to match an \textit{in-KB} mention to its gold-standard directed edges in a KB and then test on \textit{out-of-KB} mentions, following the unsupervised setting. The model architecture includes edge candidate generation with an optional step of edge selection. For edge candidate generation, we adapt the bi-encoder  \cite{wu2020blink}, with the input of a contextual mention and an edge  (i.e., edge-bi-encoder), to match a contextual mention to a directed edge in an ontology using their concept names\footnote{An edge is represented as ``parent tokens \texttt{[P-TAG]} child tokens \texttt{[C-TAG]}''.}. Top-$k$ edges rankings are selected after this step. For an optional edge selection among the top-$k$, we test the capability of zero-shot prompting of an LLM, GPT-3.5 (``gpt-3.5-turbo''), where $k$ is set as 50. The prompt includes a header, the mention with contexts, and the top-$k$ candidate edges to query the LLM to select the correct edges\footnote{Further details, parameter settings, and prompts of the experiments are available at \url{https://github.com/KRR-Oxford/OET}.}.
	
	\subsubsection*{Results on Concept Placement} 
	Table \ref{insertion-res} suggests that concept placement as edge prediction is very challenging. Also, using GPT-3.5 to select top-1 from the top-50 edge candidates does not improve, or only improves marginally, the results with the prompts. This may suggest the limitation of the state-of-the-art LLM interacting with formal, domain-specific knowledge using zero-shot prompting. 
	
	\section{CONCLUSION AND FUTURE STUDIES}
	This work introduced a new benchmark for Ontology Enrichment from Texts by Concept Discovery and Placement. The dataset focuses on enriching OWL ontologies as formal KBs, which are reducible to and thus compatible with taxonomies. Compared to the prior art, the dataset supports a more comprehensive set of characteristics, including NIL Discovery, Contextual Term, Concept Placement, and Complex Concepts. We propose a pipeline to construct this resource and release a dataset using MedMentions corpus (PubMed abstracts), UMLS and SNOMED CT ontologies. We provide usage of the data by evaluating recent LLM-based methods. 
	
	The data construction method can be applied to other KBs in the biomedical domain and KBs in various domains. 
	
	The baseline LLM-based methods are yet to achieve satisfying performance on the benchmark. Further methods are encouraged to address this challenge. 
	
	\noindent\textbf{Acknowledgements}. This work is supported by EPSRC projects, including ConCur (EP/V050869/1), OASIS (EP/S032347/1), UK FIRES (EP/S019111/1); and Samsung Research UK (SRUK).
	
	\bibliographystyle{ACM-Reference-Format}
	\bibliography{insertion}
\end{document}